\documentclass[12pt,letterpaper]{article}
\usepackage[a4paper, total={7in, 10in}]{geometry}

\usepackage[utf8]{inputenc}
\usepackage[T1]{fontenc}
\usepackage[polish,english]{babel}
\usepackage{graphicx}
\usepackage{placeins}
\usepackage[font=small]{caption}
\usepackage{helvet}
\usepackage{authblk}
\usepackage[hidelinks]{hyperref}
\usepackage{amsmath} 
\usepackage{amssymb} 
\usepackage{orcidlink} 
\usepackage[super,comma,sort&compress]{natbib}
\usepackage{cleveref}

\makeatletter
\renewcommand{\maketitle}{\bgroup\setlength{\parindent}{0pt}
\begin{center}
  \Large{\textbf{\@title}}
\end{center}
\begin{flushleft}
  \@author
\end{flushleft}\egroup}
\makeatother

\title{Remapping and navigation of an embedding space via error minimization: a fundamental organizational principle of cognition in natural and artificial systems}
\date{}

\author[1,$\dagger$\orcidlink{0000-0001-7787-4839}]{Benedikt Hartl}
\author[1,$\dagger$\orcidlink{0000-0001-8081-1070}]{Léo Pio-Lopez}
\author[1\orcidlink{0000-0002-4812-0744}]{Chris Fields}
\author[1,2,\orcidlink{0000-0001-7292-8084},*]{Michael Levin}

\affil[1]{Allen Discovery Center at Tufts University, Medford, MA, USA}
\affil[2]{Wyss Institute for Biologically Inspired Engineering at Harvard University, Boston, MA, USA}
\affil[$\dagger$]{Authors contributed equally.}

\affil[*]{Corresponding Author: \href{mailto:michael.levin@tufts.edu}{michael.levin@tufts.edu}}

\begin{document}
%%%%%%%%%%%%%%%%%%%%%%%%%%%%%%%%%%%%%%%%%%%
%%% FRONT MATTER %%%%%%%%%%%%%%%%%%%%%%%%%%
\maketitle


% \textbf{Color-Coding:}
% \begin{itemize}
%     \item {\color{teal} Ben Hartl}
%     \item {\color{olive} Léo Pio-Lopez}
%     \item {\color{blue} Chris Fields}
%     \item {\color{orange} Michael Levin}
% \end{itemize}

\colorlet{teal}{black}
\colorlet{olive}{black}
\colorlet{blue}{black}
\colorlet{orange}{black}

\textbf{Running Title:}
{\color{teal}Remapping and} Navigating Embedding Space\\

\newpage
\begin{abstract}
The emerging field of diverse intelligence seeks an integrated view of problem-solving in agents of very different provenance, composition, and substrates. From subcellular chemical networks to swarms of organisms, and across evolved, engineered, and chimeric systems, it is hypothesized that scale-invariant principles of decision-making can be discovered.  
%Here, we explore the use of a specific formalism – navigation of an embedding space – to highlight parallels between  @@@@. 
We propose that cognition in both natural and synthetic systems can be characterized and understood by the interplay between two equally important invariants: (i) the remapping of embedding spaces, and (ii) the navigation within these spaces. Biological collectives, from single cells to entire organisms (and beyond), remap transcriptional, morphological, physiological, or 3D spaces to maintain homeostasis and regenerate structure, while navigating these spaces through distributed error correction. Modern Artificial Intelligence (AI) systems, including transformers, diffusion models, and neural cellular automata enact analogous processes by remapping data into latent embeddings and refining them iteratively through contextualization. We argue that this dual principle -- remapping and navigation of embedding spaces via iterative error minimization -- constitutes a substrate-independent invariant of cognition. Recognizing this shared mechanism not only illuminates deep parallels between living systems and artificial models, but also provides a unifying framework for engineering adaptive intelligence across scales.
\end{abstract}

%%% FRONT MATTER %%%%%%%%%%%%%%%%%%%%%%%%%%
%%%%%%%%%%%%%%%%%%%%%%%%%%%%%%%%%%%%%%%%%%%

%%%%%%%%%%%%%%%%%%%%%%%%%%%%%%%%%%%%%%%%%%%
%%% OUTLINE %%%%%%%%%%%%%%%%%%%%%%%%%%%%%%%
%\newpage
%{\color{gray}\include{outline}}
%%% OUTLINE %%%%%%%%%%%%%%%%%%%%%%%%%%%%%%%
%%%%%%%%%%%%%%%%%%%%%%%%%%%%%%%%%%%%%%%%%%%

%%%%%%%%%%%%%%%%%%%%%%%%%%%%%%%%%%%%%%%%%%%
%%% CONTENT %%%%%%%%%%%%%%%%%%%%%%%%%%%%%%%

\section{Introduction}
Conventional behavioral science tends to focus on animals with brains and nervous systems. Three main lines of thought suggest that this might be too narrow, as schematically illustrated in \Cref{fig:1}. First, an operationalist stance suggests that cognitive terms are interaction protocols, not claims about the essential nature of specific substrates such as brains~\cite{Levin2022TAME, fields2022competency, fields2020scale, Lyon2021ReframingCognition, Abramson2021Behaviorist, pezzulo2016top, Lyon2015CognitiveCell, Keijzer2013WhatNervousSystemsDo, maturana2012autopoiesis, pattee2012cell, Ashby1952, james1890principles}. In other words, an assessment of cognition within any system is, operationally, just a testable hypothesis about whether, for example, the system is most amenable to techniques that rewire its hardware, reset its homeostatic goals, train it via rewards and punishments, or convince it via cogent reasons and psychological strategies. From that perspective, a system’s position on the spectrum of persuadability \cite{Levin2022TAME} is an empirical question.  This operational view facilitates the application of tools from the behavioral scientist’s toolbox \cite{Abramson2021Behaviorist} to a wide range of complex, simple, and even non-living systems to see which formal interaction model offers the most effective prediction and control. The second line of thought points out that the mechanisms implementing intelligence in the brain are evolutionarily homologous to those in the non-neural living substrates. Ion channels, electrical synapses, synaptic machinery, and the algorithms they embody are evolutionarily ancient, and ubiquitous, raising the question of when precisely they became fodder for cognitive tasks (albeit in unconventional contexts), in the long history prior to the formation of brains \cite{Keijzer2013WhatNervousSystemsDo}.  Finally, third, developmental biology reveals that familiar bioelectric networks play key roles in coordinating cellular activities toward specific anatomical endpoints, long before nerves appear during morphogenesis, again pointing to the need for models of scaling of competencies from those of the single blastomere to those of an adult \cite{levin2021bioelectric}; examples are illustrated in \Cref{fig:1}~(A). 

The emerging field of Diverse Intelligence \cite{Baluska2021BiomolecularBasis, Levin2021UncoveringCognitiveSimilarities, Lyon2015CognitiveCell, Lyon2020MinimalCognition, Lyon2021ReframingCognition} seeks to understand what all active agents have in common. In moving away from classic assumptions that brains, and a history of evolutionary trial-and-error for behavior in 3D space, are privileged substrates and origin stories for cognition, it seeks a unifying theory for agents of highly diverse composition and provenance \cite{fields2022competency}. Evolved, engineered, and hybrid systems at all scales of complexity are fodder for experiments in learning, problem-solving, and decision-making \cite{Fields2021MinimalPhysicalism, fields2020scale}. Progress in this field suggests that neuroscience is not about neurons – it’s about biophysical processes and mechanisms that operate as a kind of “cognitive glue”, binding competent subunits (e.g., cells) into collective intelligences that operate in spaces and with spatio-temporal goal horizons (cognitive light cones) much larger than those of their parts \cite{Levin2023BioelectricCognitiveGlue}.  

This research program has three unique key features. It is deeply interdisciplinary – revealing novel phenomena at the intersection of biology, computer science, materials science, and cognitive science. It is tractable – the tools of behavioral, computational, and cognitive neuroscience have now been applied far outside the brain and avail the modern worker with very effective tools with which to probe the primitive -- or perhaps surprisingly sophisticated -- minds of unconventional systems. And finally -- it impacts a critical unmet need and knowledge gap: aspects of regenerative medicine, bioengineering, AI, and even fields such as ethics are being enriched by a novel understanding of the origin, scaling, and properties of unconventional embodied minds, such as cellular collectives, that can be coaxed to pursue alternative end-states \cite{Clawson2022EndlessForms, Davies2023SyntheticMorphology, lagasse2023future, Levin2023BioelectricCognitiveGlue, Levin2024MultiscaleWisdom}. 

% {\color{orange} Here, we present a contribution to this field – a set of ideas from the field of @, which provide a new perspective on intelligence in systems that is agnostic to their scale, problem-space, or material composition. @@}

This paper extends the Fields-Levin framework of cognition as competent navigation in arbitrary spaces~\cite{fields2022competency, fields2020scale} (see  \Cref{fig:1}~(B, C)) by proposing that intelligence is characterized by two equally important cognitive invariants: the creation and \textit{remapping} of embedding spaces, and the \textit{navigation} within these spaces in response to new contexts. {\color{teal} More precisely, we define {\em remapping} in this paper as representing and/or updating new information acquired via experience into the internal latent representation formalized by the problem space (i.e., constantly refining a navigable embedding space) -- a concept closely related to internal representations formed by world-models~\cite{ha2018worldmodels, Craik1967NatureOfExplanation}.}  We argue that embedding spaces are not static and given structures but actively constructed representations that translate complex realities into navigable dimensions, as schematically illustrated by \Cref{fig:1}~(D). By recognizing both processes as fundamental to cognition, we develop a more comprehensive substrate-independent framework for understanding intelligence across biological and artificial systems. We demonstrate the explanatory power of this dual framework through examples ranging from biological regeneration to AI systems, showing how the interplay between remapping (via embeddings) and navigation processes enables complex adaptive behaviors across diverse embodiments of distributed intelligences, agnostic to problem spaces, scale, or material composition.
We begin by outlining navigation as an invariant of cognition, before turning to remapping, embeddings, and their manifestations in biology and AI. We conclude with a perspective on how these principles will foster the development of novel adaptable synthetic and chimeric collective intelligences.

\section{Remapping and navigating arbitrary spaces as an invariant of cognition}

\subsection{Navigation in arbitrary spaces as an invariant of cognition}

\begin{figure}[h!]
    \centering
    \includegraphics[width=\linewidth]{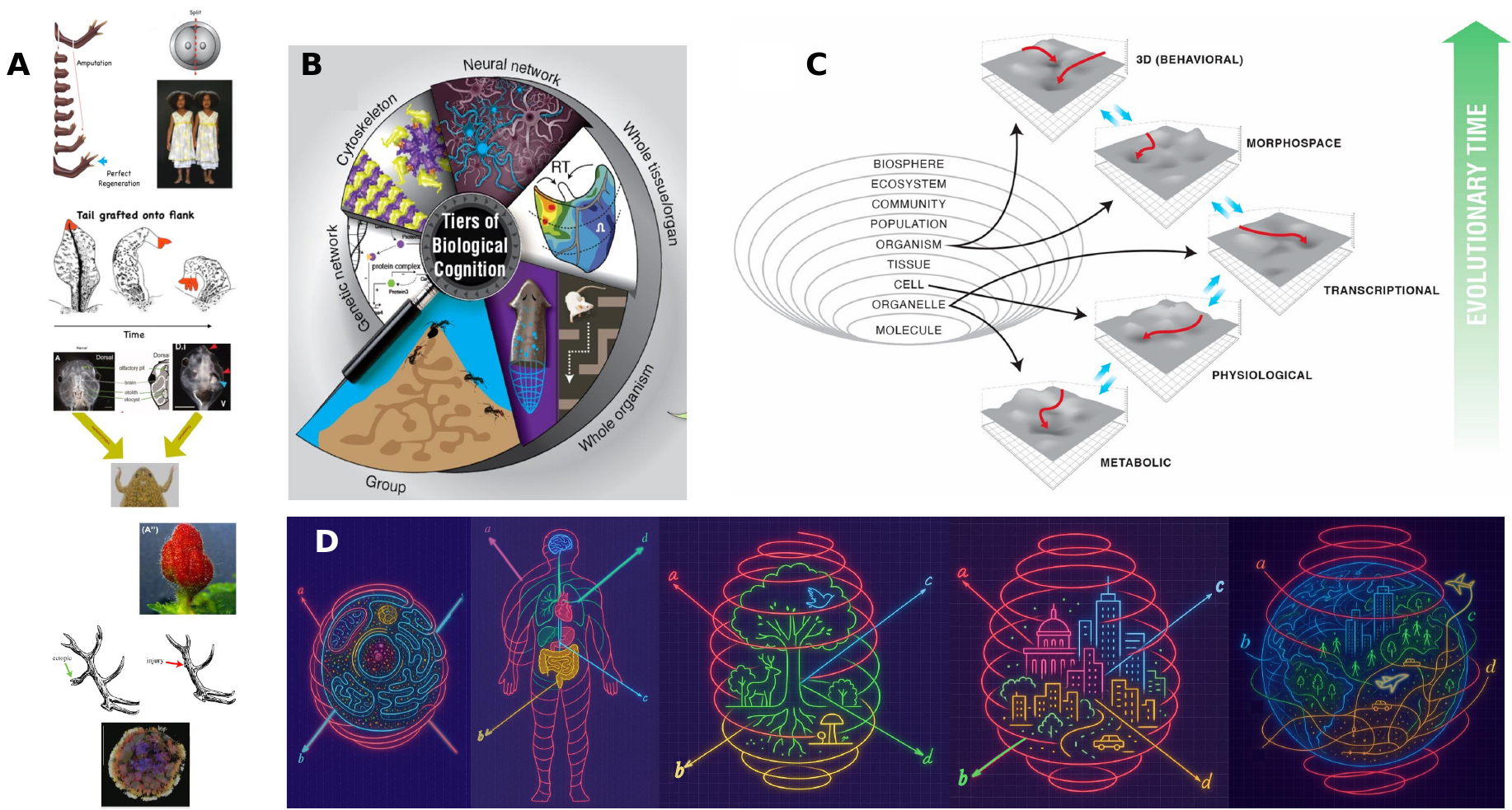}
    \caption{
        \textbf{Biology as Multiscale Competency Architecture (MCA)}:
        (A) Biological examples of unconventional morphogenetic problem solving (from top to bottom): 
        the cellular collective of regenerative species (such as salamander or axolotl) can regrow limbs, bodyparts, or organs upon injury (image by Jeremy Guay of Peregrine Creative used by permission from~\cite{levin2023darwin});
        identical twins independently develop from one zygote that has split into two embryos at an early cleavage stage (photo by Oudeschool via Wikimedia Commons);
        remodeling of a transplanted tail into a limb-like structure on a salamander's flank (used by permission from~\cite{Levin2024SelfImprovisingMemories});
        tadpoles in which the craniofacial structures are scrambled still make largely normal frogs (image taken by permission from~\cite{Vandenberg2012}, and courtesy of Erin Switzer);
        a archetypical white oak leaf (photographed by Chris Evans, River to River CWMA, \url{bugwood.org} contrasted to a novel gall form made by genetically normal plant leaf cells when prompted by signals from a wasp embryo (adapted from~\cite{Levin2024SelfImprovisingMemories});
        an injury (red arrow) altered the target morphology of a Siberian wapiti antler, producing during the following two years a new tine (green arrow), used by permission from~\cite{lobo2014linear};
        Anthrobots are living robots which spontaneously self-organize from human tracheal epithelial cells into motile, multicellular structures capable of performing useful biological tasks such as assisting neural wound healing (image from~\cite{Gumuskaya2023}).
        (B) To explain this immense structural and functional plastictiy of biological matter, collective biological systems ought to be seen as an MCA: composites of integrated layers within layers of biological self-organization, bridging spatial and temporal scales from the molecular level, over cells, tissues, organs, to whole organisms and groups or collectives of individuals. Image by Jeremy Guay of Peregrine Creative; taken from~\cite{Levin2021LifeDeath}.
        % {\color{red} possible alternatives: (i) Bio-Hybrid Fig. 2 from \cite{Clawson2022EndlessForms}, or cognitive light-cone, or newer stuff from Mike.}
        (C) A proposed model in which evolution pivots the same strategies (and some of the same problem-solving mechanisms) to navigate different problem spaces. The components of each level organize to solve tasks in their own space, and systems evolved from navigating the metabolic, physiological, transcriptional, anatomical, up to traditional behavior in 3D space, and beyond; taken and adapted from \cite{fields2022competency}.
        (D) Schematic of multiscale embedding architectures -- from cells to organisms, ecologicosystems, cities, and planetary-scale systems -- illustrating that each level constructs and operates within its own embedding space while simultaneously shaping those of the levels above and below.
        Across scales, cognition can be understood as the invariant capacity to navigate and remap these embedding spaces in response to changing internal or external conditions.
        Panel D is an AI-generated schematic illustration created using ChatGPT (OpenAI) based on author-provided text prompts inspired by~\cite{Ananthaswamy2023NewApproach}.
    }
    \label{fig:1}
\end{figure}

William James, the pioneering psychologist, offered a goal-directed definition of intelligence: ``Intelligence is a fixed goal with variable means of achieving it'' \cite{james1890principles}. It remains particularly prescient in the modern era of AI, where populations of software and physical agents are expanding. This definition also emphasizes that intelligence manifests as goal-directed behavior rather than any specific mechanism or embodiment. The cybernetic tradition, developed by Ashby \cite{Ashby1952}, Maturana and Varela \cite{maturana2012autopoiesis}, and Pattee \cite{pattee2012cell}, expanded this perspective by framing cognition in terms of information processing and self-regulation. Very recently, Fields and Levin expanded further this approach and framed cognition as competent navigation in arbitrary spaces \cite{fields2022competency}, as illustrated in \cref{fig:1}~(C). This framework proposes navigation in a problem space (metabolic, physiological, morphological, or behavioral for biological systems, for example) as an invariant for analyzing cognition in diverse embodiments \cite{fields2022competency}. More specifically, in this framework, a goal is a (target) position in a problem space, while a system's degree of competency (or intelligence, in the definition of William James) is reflected by the system's ability to navigate that space toward a goal state by actively minimizing the deviation from diverse starting positions. Such a deliberately substrate-agnostic framework allows recognizing common patterns of intelligence across vastly different systems \cite{levin2023darwin}, from biological to artificial systems.

There are several examples of navigation in spaces in biology, including the metabolic, physiological, transcriptional, morphological or behavioral (3D) spaces. Biological systems are the result of multiscale goal-directed behavior (see \Cref{fig:1}~(A, B)). These navigational capabilities are indeed not isolated but interconnected through hierarchical relationships that enable coordination across scales for reaching homeostasis~\cite{levin2022technological,fields2020scale}. A key insight of the Fields-Levin framework is how higher-level systems influence lower-level ones (see \Cref{fig:1}~(C, D)). Higher-level systems deform the energy landscapes for their subsystems, allowing lower-level components to follow local gradients that achieve goals beneficial to the higher-level system \cite{fields2022competency}. This allows the scaling of goals and the development of higher cognition \cite{pio2023scaling} and creates a situation where components can engage in relatively simple local behaviors while collectively achieving sophisticated system-level outcomes.  

% Another competent navigation achieved by biological non-neural systems happens in the morphospace 
Intelligence in animals is typically measured by their capabilities to solve problems by navigating and manipulating objects, and resolving problems in 3D space \cite{MacLean2014EvoSelfControl, Seed2010AnimalToolUse, shettleworth2009cognition, Emery2004MentalCrows, okeefe1978hippocampus, Tolman1948CognitiveMaps}. However, it becomes increasingly obvious that certain (different) levels of agency and cognition need to be attributed to the components of all organizational layers across the multiscale competency architecture of biology~\cite{levin2023darwin} (see \Cref{fig:1}~(B)), Even cells and cellular collectives solve problems when navigating transcriptomic space or morphospace -- the space of possible anatomical configurations \cite{levin2022technological} -- which leads to the immense structural and functional plasticity of living matter; selected examples are visualized in \Cref{fig:1}~(A) and discussed below. 

%%% PLANARIA BARIUM
The planarian flatworm provides an interesting example of intelligent navigation in the transcriptional space \cite{levin2019planarian}. Exposure to barium, a potassium channel blocker, causes head degeneration. Within the same specimen, planaria regenerate new heads with the transcriptional changes needed to achieve barium resistance \cite{durant2017long}. How can groups of cells identify which of tens of thousands of potential transcriptional effectors will solve their physiological stressor? Moreover, barium is not something planaria encountered in the wild, suggesting cells possess general problem-solving capabilities, not only genetically pre-programmed responses to specific challenges \cite{levin2022technological}. The system senses its current position (a disrupted physiological state), compares it to a preferred position (normal function), and moves through transcriptional space by modifying gene expression until it reaches a new attractor region that restores homeostasis. There is not enough time to do this by trial and error, and some sort of heuristic must be used to identify solutions to novel problems in this (and other) spaces that life thrives in. Even below cell level, gene regulatory networks are capable of various kinds of learning and memory \cite{biswas2021gene, biswas2022learning}.

During development and regeneration, cellular collectives navigate morphospace with high robustness, coordinating their activities to achieve specific target morphologies despite varying starting conditions or perturbations \cite{levin2022technological,levin2021bioelectric}.  Salamanders can regenerate limbs from arbitrary amputation planes \cite{tanaka2016salamander}, cellular collectives in a tadpole's disrupted ``Picasso face'' reorganize to form normal frog facial structures \cite{vandenberg2012normalized}, and planaria can regenerate heads with species-specific morphologies without genomic editing \cite{levin2012morphogenetic, durant2017long}. These processes may involve cellular collectives actively sensing their current morphological state, comparing it to the target morphology, and executing collective cellular decision-making to reduce the error, probably using biochemical, mechanical and bioelectrical cues \cite{levin2022technological}.  The cellular collective constructs a distributed representation of both its current and target morphology, then navigates through morphospace using local cell behaviors that collectively produce global pattern formation \cite{wolpert2015principles}. The Fields-Levin framework defines these phenomena as navigation in morphospace toward specific attractor regions \cite{fields2022competency} (c.f., \Cref{fig:1}~(C)). 

This framework is not restricted to cells that form tissues or morphologies, but scales across all organizational layers of biology: distributed agents perceive and act locally, following their individual agenda that is aligned with, or constrained by, a collective higher-level objective.
Moreover, navigation in a problem space is not limited to biological systems, Large Language Models (LLMs) navigate the language space \cite{rogers2020primer,yang2024harnessing}, robots navigate motor spaces \cite{schaal2006dynamic}, multimodal models are navigating a merged space of different modalities \cite{baltruvsaitis2018multimodal}, etc.

% I've moved/repurposed this as last paragraph of the intro
% This paper extends the Fields-Levin framework of cognition as competent navigation in arbitrary spaces by proposing that intelligence is characterized by two equally important cognitive invariants: the creation and remapping of embedding spaces, and the navigation within these spaces. We argue that embedding spaces are not static and given structures but actively constructed representations that translate complex realities into navigable dimensions. 
% By recognizing both processes as fundamental to cognition, we develop a more comprehensive substrate-independent framework for understanding intelligence across biological and artificial systems. We demonstrate the explanatory power of this dual framework through examples ranging from biological regeneration to artificial intelligence systems, showing how the interplay between remapping (via embeddings) and navigation processes enables complex adaptive behaviors across diverse embodiments.

% I've tried to make a brief transition motivating the next section
Navigation in arbitrary spaces, from morphogenesis to generative AI, reveals a common substrate-independent principle of cognition. Crucially, however, the state spaces are not static backdrops: they are continually constructed and reshaped by the systems that use them under novel contexts. Through this lens, remapping becomes inseparable from navigation, and together they constitute dual invariants of intelligence, as we turn to below.

\subsection{Navigation implies remapping for learning systems}

As we just saw, competent navigation in a problem space is a hallmark of cognition in the Fields-Levin framework~\cite{fields2020scale, fields2022competency}. However, any learning system must remap and update this problem space when it acquires new knowledge on the internal and external worlds. We believe remapping the problem space cannot be separated from competent navigation in this space to define cognition and intelligence. %More precisely, we define {\em remapping} in this paper as representing and/or updating new information acquired via experience into the internal representation formalized by the problem space -- a concept closely related to internal representations formed by world-models~\cite{ha2018worldmodels, Craik1967NatureOfExplanation}.  
Remapping, according to {\color{teal} the above of} definition {\color{teal} -- as representing and/or updating new information acquired via experience into the internal representation formalized by the problem space --} combines learning with experience-dependent modification of the state space in which learning takes place.  In biological systems, remapping occurs at several levels. Neural plasticity, for example, represents a remapping of the brain's memories and representational space, where connections and weightings are reconfigured based on experience, effectively changing not only what is thought, but also the landscape of possible thought patterns \cite{kandel2001molecular, martin2000synaptic}. Specific neural structures do the same, for example, the hippocampus demonstrates this kind of remapping for navigation in different spaces, creating new internal spatial representations when animals encounter novel environments \cite{fenton2024remapping}. But remapping is not only limited to the brain -- the adaptive immune system actively remaps its recognition spaces using VDJ recombination, creating new antibody configurations that represent previously unencountered antigens \cite{tonegawa1983somatic,schatz2011v}. The already described planarian's transcriptional adaptation to barium demonstrates remapping of its internal transcriptional landscape to represent and respond to this novel environmental chemical \cite{durant2017long}. At cellular levels, stem cells while differentiating are remapping completely their transcriptional landscape too \cite{laurenti2018haematopoietic}. Even molecular networks can learn and have different types of memory, actively updating and remapping their internal representations of the learned information \cite{biswas2021gene, biswas2022learning}.

Sensorimotor systems provide particularly clear examples of remapping via the creation of internal representations by building neural maps. These maps are found ubiquitously in the brain, where they encode a wide range of behaviorally relevant features into neural space. Neural maps in the brain exhibit different organizational patterns. Continuous maps maintain spatial relationships where neighboring neurons encode similar stimulus properties. For example, retinotopic maps in visual processing areas preserve spatial relationships from the retina \cite{benson2012retinotopic}, and neurons in tonotopic maps are organized following sound frequency gradients, with neighboring cells responding to similar frequencies \cite{moerel2014human}. Discrete maps, by contrast, create distinct functional modules without gradual transitions. The olfactory system is one of them: olfactory receptors located throughout the nasal cavity connect to specific neural clusters called glomeruli in the olfactory bulb, with each receptor type linking to its own dedicated glomerular group \cite{nakashima2024roles}. Many brain regions combine both organizational strategies, for example somatosensory areas processing touch and regions controlling movement typically integrate continuous gradients with discrete frontiers, creating hybrid mapping systems that balance precision with flexibility within two-dimensional cortical surfaces \cite{luo2007development}. Critical period plasticity is also a remapping, where sensory maps are reconstructed to represent altered input patterns during development \cite{hensch2005critical}. 

The evolutionary perspective reveals remapping as fundamental to adaptation. In fact, as our senses have evolved, we have seen new maps for translating external energy—such as light, sound, and chemicals—into internal neural representations. The development of bat echolocation, for instance, necessitated remapping auditory processing in order to produce intricate internal spatial maps from external acoustic reflections \cite{griffin1958listening}. These sensory maps are continually being updated and revised. Primary sensory cortices constantly remap the organization of their receptive fields to reflect changes in the environment \cite{gilbert1992receptive}. Cross-modal plasticity after sensory loss shows a drastic remapping. For example, in blind  \textcolor{olive}{subjects}, the visual cortex literally represents external auditory information in the previously visual neural cortices by creating new internal maps of auditory space in these cortical areas \cite{strelnikov2013visual}. When learning new primitives and movement pattern sequences, or when internalizing external tool representations, the motor system also exhibits remapping. Indeed, the motor cortex develops new maps that represent the relationship between intended actions and their effects in the external world \cite{kalaska2009intention, harrison2014motor}. Maybe even more remarkable, brain-machine interfaces demonstrate remapping in real-time, as motor cortex represents external device control, mapping internal neural activity onto external robotic movements \cite{lebedev2006brain}.

\begin{figure}
    \centering
    \includegraphics[width=\linewidth]{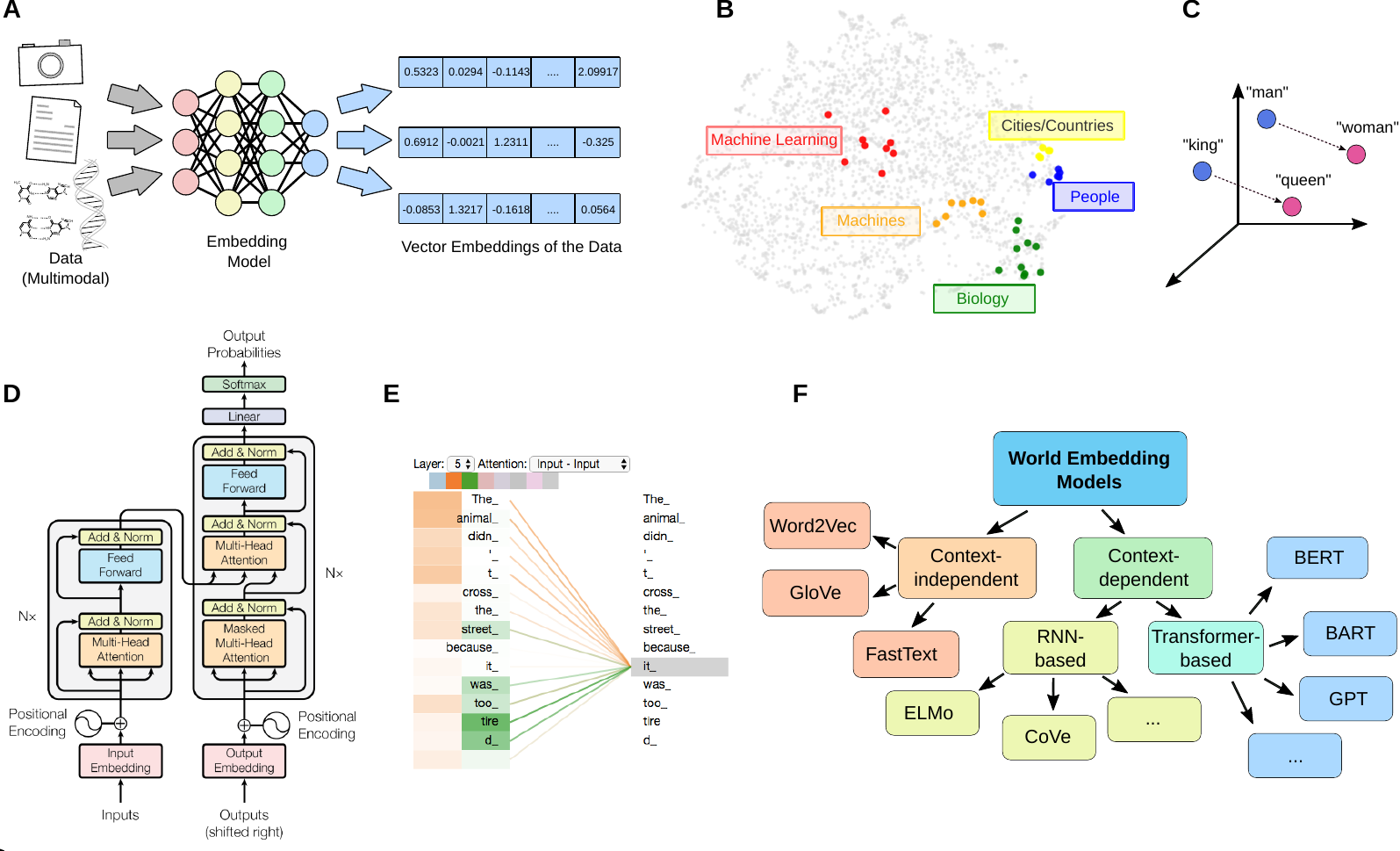}
    \caption{
        \textbf{Data-embeddings and transformers:}
        (A) Embedding models transform multimodal data into standardized numerical representations, i.e., low-dimensional vectors in an abstract embedding space (compared to the high-dimensional unstructured data), where directions encode semantic relations.
        (B) Low-dimensional projection of word-embedding vectors of the content of the present manuscript, showing clusters of semantically related words (color-coded).
        (C) Geometric relations in embedding spaces capture semantic analogies, e.g., the vector offset "man" $\rightarrow$ "woman" parallels "king" $\rightarrow$ "queen" in Word2Vec space~\cite{mikolov2013Word2Vec}.
        (D) The transformer architecture forms the backbone of modern language and vision models. Transformers interpret their stream of input data (i.e., the context) by computing attention scores between all input tokens (e.g., word or image-patch embeddings), allowing the model to weigh the tokens' relative importance. Image adapted from \cite{vaswani2017attention}.   
        (E) Visualization of the attention mechanism in action during next-token prediction, illustrating how a model distributes attention across tokens when predicting the next word. As the model encodes the word ``it", different attention heads focus on different parts of the context -- for example, one head attends primarily to ``the animal" while another attends to ``tired". Thus, the model’s representation of ``it" incorporates information from both semantic sources, illustrating how attention mixes contextual representations in associative embedding space~\cite{ramsauer2021hopfield} (image adapted from \cite{jalammar2016IllustratedTransformer}).
        (F) An overview of word embedding models. 
        Word embedding approaches can be grouped into context-independent and context-dependent representations.
        Context-independent models assign each word a single fixed vector regardless of usage (e.g., Word2Vec~\cite{mikolov2013Word2Vec}, GloVe~\cite{pennington2014glove}, FastText~\cite{bojanowski2017enriching}).
        Context-dependent models generate token embeddings that vary with linguistic context, either through recurrent neural networks (ELMo~\cite{Peters2018ELMo}, CoVe~\cite{McCannCoVe2017}, etc.) or transformer architectures (BERT~\cite{devlin2019bert}, BART~\cite{lewis2020bart}, GPT~\cite{radford2018improving}, etc.) -- and thereby marked a paradigm-shift in embedding models.
        Arrows indicate conceptual or historical advances.
        The overview is adapted from~\cite{spotintelligence2023embeddings}.
    }
    \label{fig:data-embeddings}
\end{figure}

% Remapping also exists in artificial systems.
Just as biological systems construct and revise internal maps to maintain homeostasis and adapt to novel challenges, AI systems build and remap data into structured spaces, via embeddings. These parallels suggest that remapping is a general strategy for intelligence, irrespective of substrate. We now turn to AI systems to illustrate this convergence. LLMs for instance, generate embeddings on a vector space that are constantly updated with new data (c.f., \Cref{fig:data-embeddings}). Embeddings are vector representations of tokens, and distances and directions between them represent semantic relationships in the text \cite{brown2020language, dong2022survey}. Computer vision systems transform visual scenes into hierarchical maps that show external objects together with their textures and spatial connections \cite{lecun2015deep}. The internal policy maps of reinforcement learning systems demonstrate how external states connect to optimal actions \cite{sutton1998reinforcement, mnih2015human}. Transformers' attention algorithms generate internal maps of token relevance which represent the external context structure that matters for each processing step \cite{vaswani2017attention}. Generative models in AI transform external data distributions into internal latent spaces to create compressed representations of external patterns \cite{kingma2013auto,goodfellow2014generative}. Robots produce actual navigation maps when they explore new environments \cite{filliat2003map}. Few-shot learning models exhibit fast remapping capabilities through their ability to develop new internal representations that connect new classes to existing representational spaces \cite{finn2017model}. Meta-learning systems possess advanced remapping abilities which allow them to establish new internal maps for external tasks by choosing appropriate learning methods \cite{lake2017building}.

Through remapping at all levels, learning systems can discover new representational schemes that capture previously unrecognized external patterns, create internal maps that efficiently compress external complexity, and integrate information from heterogeneous external sources into unified internal maps/spaces \cite{epstein2017cognitive, alvernhe2011local, he2019manipulating, schuck2019sequential, garvert2017map}. By recognizing remapping as the fundamental process of competently navigating a problem space, we develop a more precise substrate-independent framework for understanding intelligence across biological and artificial systems \cite{fields2022competency,levin2022technological}.

\section{A common mechanism between AI and biological systems? Remapping via embeddings.}

The kinds of remapping observed in biological systems and employed in AI systems are not arbitrary, but rather appear tightly controlled.  What constrains remappings, and how is this constraint information encoded by the system doing the remapping?  We suggest here that coarse-graining {\em embeddings} provide a general mechanism for constraining remappings, one that can enforce strong constraints without representing them explicitly.  Embedding a high-dimensional parent space $\Gamma$ into a lower-dimensional latent space $\Xi$ via a structure-preserving (i.e. non-trivial) map $\xi: \Gamma \hookrightarrow \Xi$ selects, in particular, parent-space structures or relations that are inverse images, under $\xi$, of structures or relations in the smaller latent space (c.f., schmatic illustration in \Cref{fig:1}~(C, D)).  We first examine embeddings of biological systems into 3D space, then consider more general embeddings available to AI systems.

\subsection{Embedding in 3D space as a remapping constraint}

Biological systems continually traverse metabolic and gene-expression (i.e. transcriptomic and proteomic) spaces, and in doing so, sometimes also explore morphological space.  These spaces can be very high-dimensional, involving tens to hundreds of thousands of chemical potentials, each with dynamic ranges of at least an order of magnitude.  The vast majority of states in these biochemical state spaces are not just far from homeostatic ranges but inconsistent with life, e.g. states in which most proteins are denatured or states in which essential enzymes are deficient.  A latent space specific to living systems, therefore, must encode a set of constraining relations that appears to be combinatorially explosive.  If we knew this latent space and the embedding map to it, biochemistry would be finished.

Nature does not, however, have to know either the latent space or the embedding.  Nature embeds biochemistry in 3D space, and selects for embedding maps from chemical-potential space to bounded volumes in 3D space with a simple Galilean metric.  Almost all of these maps do not work, but some do.  The maps that work implicitly encode constraints on molecular relations that are consistent with life.   Some of these constraints are extremely tight, forcing some pairs of molecules to be and remain in close proximity in 3D space.  Others are much looser, allowing molecules to diffuse through the highly-structured 3D space specified by the tight constraints.  We, as biochemists, try to figure out those implicit constraints by doing experiments on the systems that we observe, i.e. those that ``work'' in 3D.  
    
Embedding biochemistry in 3D converts navigation in chemical-potential space into molecular behavior in 3D space, with the constrains imposed by the embedding map manifested as behavioral constraints on the embedded molecules.  Protein folding, for example, is highly constrained by ionic and van der Waals interactions that are defined in 3D space, not in the parent concentration space.  Binding pockets, hydrophobic interactions, and diffusion are, similarly, spatial concepts that express navigational biases when pulled back to the parent space.

The idea of embeddings as imposing constraints can be made precise in the language of sheaf theory.  Sheaves are a category-theoretic description of what is common to, for example, vector fields, maps or geographic information systems, databases, and other coherent ways of ``attaching'' structured data to each point or ``location'' in some space \cite{hartshorne, spivak}; see Supplementary Information for formal definitions.  Coherence is enforced by requiring, roughly speaking, that assignments of data to different regions agree wherever the regions overlap, and that any way of building up the data for a region by assigning data to subregions agrees with every other way of building up the data for that region by assigning data to subregions.  Letting the ``space'' to which data are attached be 3D space extended along a time axis, i.e. a 4D spacetime, a coherent embedding of biochemistry is one in which every state and state change can be implemented by a time-evolving 3D structure in a way that respects all local 3D constraints, e.g. the structures of other nearby molecules.  Note that coherence in this sense is a necessary, but not sufficient, criterion for viability.  Both the biochemistry of live cells and that of lysed, dying cells must embed in 4D.

Probability distributions are assignments of data -- probabilities -- to underlying data sets or spaces.  Violations of the coherence conditions correspond to violations of the Kolmogorov axioms, e.g. local additivity of probabilities.  Abramsky and Branderburger \cite{abramsky} have shown that data violating the sheaf coherence conditions can be viewed as outcomes of measurements, or in general, processes exhibiting noncausal context dependence, e.g. processes that exhibit quantum interference.  In the current context, cellular processes that mutually interfere will appear ``non-local'' in a 4D embedding, and will be disrupted (or ``decohered'' in quantum-theoretic language) by measurements or other processes that enforce locality \cite{fg:25}.

\subsection{Multi-scale embeddings and coherence conditions}

The foregoing says nothing about the scale of the embedding; different 4D embedding scales for biochemistry, for example, capture molecular dynamics at different resolutions \cite{zweir:10}.  Embeddings of the same data at different resolutions are coherent if, but only if, the higher-resolution embeddings coarse-grain to the lower resolution embeddings via some function that does not depend on the embedding-space coordinates.  High-resolution constraints that do not ``smear out'' at lower resolution may appear as ``emergent structures'' or ``emergent boundary conditions'' in the lower-resolution description \cite{fl:25}; for example, networks of interactions that, at high resolution, limit free diffusion may appear as domain walls or mesoscale structures at low resolution.

In the context of the Fields-Levin model \cite{fields2022competency}, spatial embeddings may extend beyond the cellular scale to those of tissues, complex multicellular bodies, or communities of interacting organisms.  These scale changes introduce ``natural'' spatial boundaries, e.g. the boundary of a tissue or an individual organism, that function as emergent boundary conditions on macroscopic behavior.  One effect of these boundary conditions is to enforce spatial symmetries by enforcing spatial coupling between micro- to mesoscale systems and their local environments; organismal boundaries, for example, keep the parts of an organism together by assuring that they are transported through space together.  Processes that would not be translationally invariant without such boundaries -- e.g. a GRN moved from inside an organism to the ambient inter-organism environment -- become effectively translationally invariant because their local environments move with them.  These boundary-imposed symmetries render ``space'' and hence the spatial embedding functionally meaningful at larger scales.

It is this preservation of effective spatial symmetries by emergent boundaries that allow processes as well as structures to be spatially embedded.  This is true not just of biochemical, e.g. metabolic or signal transduction, bioelectric, and mechanical processes, but also of cognitive processes.  Active inference \cite{friston2010}, for example, can only be maintained in a reasonably familiar and hence reasonably predictable environment; completely unfamiliar environments -- e.g. environments involving radically different chemical potentials, EM fields, particle fluxes, or mechanical forces -- tend to induce Markov blanket collapse, generative-model dysregulation, and death.  From this perspective, the usual inferences from cognition to niche construction and from more complex, e.g. human cognition to (often by analogy) less complex or basal cognition appear both counter-intuitive and unmotivated.  Small, low-complexity environments are both more probable and easier to maintain than large, high-complexity environments, and relative environmental stability is as essential for cognition as it is for metabolism.  The former, like the latter, can be expected both to start out and to be widespread in small, relatively low-complexity systems coupled tightly to and spatially transported with small, relatively low-complexity environments, the maintenance of which becomes the primary objective of active inference by the systems inhabiting them.  These conditions are met at the scales of functional networks within cells, including GRNs, single cells including prokaryotes, and communities such as microbial mats.  Coherent spatial embedding, and hence the possibility of active inference, starts small and simple, and expands to larger scales as larger-scale spatial constraints emerge.  

The principles observed in biological systems -- navigation of attractor landscapes, local error correction that scales to system-level outcomes, and nested embeddings that support higher-level organization -- exemplify scale-free cognitive dynamics where local competition and coordination give rise to the abstraction of intelligence across scales. These principles are not unique to living tissue but also apply to artificial systems, as discussed in the following.

\subsection{Remapping and Navigation of Embedding Spaces in Modern AI}
Above, we discussed fundamental organizational principles in biological systems based on dynamic, context-sensitive remapping of embedding representations. AI is undergoing a similar evolution. Traditional machine learning models, such as classifiers or auto-encoders, often acquire fixed feature vectors and latent-space representations representable as data manifolds \cite{lecun2015deep}.  These learned representations performed well for narrowly defined tasks but lacked the flexibility to adapt to new situations in broader contexts and across multimodal data.

Instead of relying on static representations, modern AI increasingly relies on generative models \cite{Foster2023generative} -- models that learn the underlying distribution of an often vast amount of training data -- to construct dynamic and context-sensitive embedding spaces. So-called foundational models trained on ``internet-scale'' data can reach remarkable generalization capabilities \cite{sutton2019bitter, yousefi2024bitterlearning} (but may lack the precision required for narrow tasks due to ``over-generalization'' \cite{Li2025}). Here, an embedding is a mapping from complex, often high-dimensional unstructured data into a structured numerical vector space -- typically still high-dimensional -- that preserves similarities and emphasizes semantic relations in the remapped representation. For example, words with similar meanings are mapped closely together, and associated sequences of words are often parallel trajectories in the embedding spaces of language models \cite{mikolov2013Word2Vec} (see \Cref{fig:data-embeddings}). 

In transformer architectures \cite{vaswani2017attention}, this is realized by an attention mechanism~\cite{Bahdanau2015Attention}. Each attention layer computes how tokens relate to one another in the current context (the current text input sequence) and updates their embeddings accordingly. Crucially, this remapping of token embeddings updates the input token sequence during the pass through the model, i.e., it does not replace the latter: Across a hierarchy of attention layers, token embeddings are thus remapped - or refined - incrementally to create increasingly contextually enriched token representations for a given input sequence. Moreover, transformers maintain several attention heads. Those are sensitive to particular modalities in the data and can thus be thought of as filtering the embedding space in parallel for meaningful relationships across certain semantic features, much like how neural circuits map multimodal sensory inputs.
This allows transformers to reason about relationships and contexts through navigation in dynamically remapped embedding spaces. Strikingly, these transformer architectures can also be applied to the vision domain \cite{Dosovitskiy2021VIT}, and the learned vision embeddings can be repurposed -- understood by \cite{Jha2025UniversalEmbedding} -- language models to form powerful multimodal embedding models \cite{radford2021CLIP}.

Recently \cite{krotov2025modernmethodsassociativememory, ramsauer2021hopfield}, the self-attention mechanism in transformers has been shown to be mathematically equivalent to a form of continuous Hopfield network. The latter can store a large -- potentially exponential -- variety of data patterns as attractor states in a high-dimensional energy landscape. Upon noisy or even novel inputs, Hopfield networks retrieve the closest stored pattern and thus act as associative memories \cite{hopfield1982} of their training data. Notably, this happens via internal energy minimization dynamics, i.e., equilibration into attractor states in high-dimensional associative embedding spaces. On the one hand this implements a form of pattern-level (global) error correction. On the other hand, the hierarchical attention layers act as a kind of iterative associative memory retrieval under a given context, selecting and refining information that best matches the input query among stored patterns.

The related concept of hyperdimensional computing \cite{kanerva2009hyperdimensional} (HDC) is built on the principle that information can be stored and manipulated in extremely high-dimensional spaces -- often 10,000+ dimensions -- where different concepts are robustly represented as orthogonal vectors across all coordinates. In particular, HDC utilizes algebraic operations to superimpose elemental concepts to construct composite representations (e.g., the vector for a \textit{red car} $\approx$ \textit{red} $\bigoplus$ \textit{car}), and even allows the storage of entire sequences as single vectors \cite{burrello2020hyperdimensional}. Because different features are spread redundantly, or "holographically" across the full vector space, superposed memories can be recovered even when a large fraction of the dimensions are corrupted. This renders HDC highly robust and akin to associative memories. Embedding models, through their high-dimensional embedding representations and attention mechanisms, may implicitly exploit similar principles: Attention layers map and combine sequences of features onto high‑dimensional vectors on the fly, enabling such models to retrieve relevant patterns iteratively -- layer by layer -- during inference and capture context‑dependent associations.

While transformers capture intricate contextual relationships through their stacked attention layers, and although their context is recursively extended by appending tokens (next token predictions), their architecture remains fixed and strictly feedforward. In turn, their internal representations are generated \textit{de novo} at every inference step. This might fundamentally affect their reasoning performance through limited cognitive depth - a matter that is fiercely debated by the community \cite{lindsey2025biologyLLMs,Zhong2025FromSystem1ToSystem2, Lawsen2025IllusionOfIllusionOfThinking, Parshin2025IllusionOfThinking, Takeshi2022LLMsZeroShotReasoners}. Emerging research explores recurrent embedding architectures that explicitly utilize internal states over time: RWKV \cite{peng2023rwkv}, Mamba \cite{gu2024mamba}, or xLSTM \cite{beck2025xlstm} architectures, to name but a few, maintain internal embedding states that capture and balance short- and long-term temporal dependencies with attention-like mechanisms. These models show potentially on-par or superior performance at a fraction of computational costs compared to transformer architectures. 

Even more ambitiously, world model architectures \cite{ha2018worldmodels, dawid2023AutonomousMachineLearning, Assran2023JEPA, assran2025vjepa2selfsupervisedvideo} aim to build models that understand the consequences of their actions in an environment. Such world model architectures maintain internal states of their environment, and actively try to minimize deviations between how their internal model of the world (expectations) and the actual perceived environment evolve, given their actions. This requires these models to build compact internal representations that capture essential features across scales of unstructured spatial and temporal input \cite{dawid2023AutonomousMachineLearning}. Such a mechanism can potentially be model-agnostic and enables planning and imagination rather than just reactive processing, potentially improving learning performance through ``in-dream'' training in several iterations of the model's own imagination \cite{ha2018worldmodels}. These models represent a next step in AI, shifting from static remapping of data to constructing internal simulations of compressed representations, or embeddings, that guide future actions. This increasingly aligns such AI systems with active inference frameworks (see below, \cite{friston2010, Friston2021worldmodel}) and biological cognitive agents \cite{leadholm2025_1000brains}. 

In general, identifying well-structured embedding spaces is critical. It has recently been shown \cite{Jha2025UniversalEmbedding} that sufficiently large embedding models that are trained on a substantially general data corpus ALL identify virtually the same uniform embedding-space geometry, {\color{teal}which is not only relevant for interfacing multimodal embedding architectures, but} raising fascinating questions about the Platonic representation hypothesis \cite{Huh2024PlatonicRepresentation, Levin2025IngressingMinds}. For smaller everyday models, representations learned via open-ended neuroevolution paradigms show superior embedding representations as opposed to gradient-based methods \cite{Kumar2025Fractured}: despite both methods resulting in similar overall task performance and training dynamics \cite{Whitelam2021}, they often fundamentally differ in their internal data representations. Without particular regularization techniques, gradient-based methods often suffer from disorganized "entangled" embedding structures, termed fractured entangled representation (FER) \cite{Kumar2025Fractured}. In contrast, neuroevolution techniques biased towards open-ended search tend to find modular "disentangled" representations that generalize well to further downstream tasks. Identifying maximally informative, disentangled, conceptually orthogonal embeddings in a systematic way is thus invaluable to advance representation learning. Promising self-supervised techniques like VICReg \cite{Bardes2022} or JEPA \cite{Assran2023JEPA, assran2025vjepa2selfsupervisedvideo} impose constraints on variance, invariance, and conceptual decorrelation by whitening covariance across latent variables, preventing collapse into trivial solutions while preserving meaningful relationships between latent space dimensions.

Across various substrates, such as cells, brains, or AI architectures, embeddings provide representational, context-sensitive abstractions, while self-regulatory error correction drives the system's dynamics. Intelligence does not emerge from recombining static maps but from iterative cycles of prediction, correction, and refinement within contextually (dynamically) remapped embeddings. We argue below that iterative error correction represents an operational backbone linking navigation and remapping.

\section{A pervasive error-correction scheme in AI and biological systems}
% \begin{itemize}
%     \item Active inference everywhere (where are the boundaries, are they fractal?) 
%     \item Diffusion and predictive coding, link with active inference/conditional diffusion models 
%     \item Diffusion models and LLMs, error correction mechanism: incremental teleology (small noisy steps towards an envisaged goal) 
%     \item error minimization everywhere (from sub-cellular decisions to financial audits), prediction, planning as inference, memory, in brief, cognition 
%     \item NCA and embedding (more flexible ANNs than DMs, distributed self-regulation across scales) 
%     \item Multi-scale competency architecture in biology 
%     \item Concurrent navigation at different scales  [ polycomputing? ] 
% \end{itemize}

From cells to human minds, adaptive agents continuously predict the state of their environment against perceptions and adjust their internal models or their actions to reduce prediction errors \cite{friston2010, clark2013, pezzulo2016top, pio2022active, fields2022competency, levin2022technological}.
This prediction--perception--error-correction loop represents a robust navigation policy in arbitrary physical, cognitive, or virtual embedding spaces \cite{fields2022competency, fields2020scale}, and has started to gain traction in machine learning research; examples are transformers incrementally refining streams of tokens using stacks of attention heads, or world models updating their internal representations based on new evidence.

While iterative self-regulatory error-correction provides scale-free dynamics for cognitive processes \cite{fields2020scale}, remapping representations under novel contexts adds plasticity to an agent's internal model, effectively facilitating creativity and learning \cite{Levin2025DiverseIntelligence, fields2022competency, pezzulo2016top, fields2020scale, friston2010, Friston2021worldmodel, levin2023darwin}. From a practical point of view, we suggest that self-regulatory error-correction in observer-specific plastic embedding spaces is a unifying operational principle for biological and artificial intelligence.

In cognitive systems, this principle is formalized through the variational free-energy principle \cite{friston2010}, or active inference \cite{friston2016activeinference, parr2022activeinference}, which infers from fundamental physics considerations \cite{ramstead:22, friston:23} that self-organizing systems must act to minimize an upper bound on surprise, i.e., on unpredictable environmental fluctuations. Whether at the scale of a single cell responding to morphogen gradients, or an organism planning future actions, systems evolve internal models that capture the latent structure of the actions of their environments on their boundaries. In the space defined by system-environment information flows, these boundaries are Markov blankets \cite{pearl:88}; embedding this information-flow space into 3D typically maps such boundaries into meso- to macroscale emergent structures, e.g., cell membranes or the skins of multicellular organisms, that spatially separate internal from external degrees of freedom and processes. Complex systems comprise nested boundaries that facilitate localized prediction and correction processes, with each component system having to cope with only a local, spatially co-transported environment as discussed above. Crucially, in the multiscale competency architecture of biology \cite{levin2023darwin} these boundaries are not static but dynamic and possibly fractal, scaling from sub-cellular compartments to entire organisms and possibly beyond \cite{Levin2025DiverseIntelligence, mcmillen2024collective}.

Modern AI approaches have adopted ideas from active inference in several ways. World models \cite{ha2018worldmodels, dawid2023AutonomousMachineLearning, Friston2021worldmodel}, for instance, embody this principle probably most explicitly by minimizing prediction error between perceptions and a self-maintained internal state predictions through belief-state updates and/or taking environmental actions.
Other related developments have managed to train large language models (LLMs) to display advanced reasoning solely through self-supervised learning \cite{zhao2025learningreasonexternalrewards}: a completely intrinsic reward mechanism maximizing the model's confidence in its outputs appears to surpass customized external reward schemes.

\begin{figure}
    \centering
    \includegraphics[width=\linewidth]{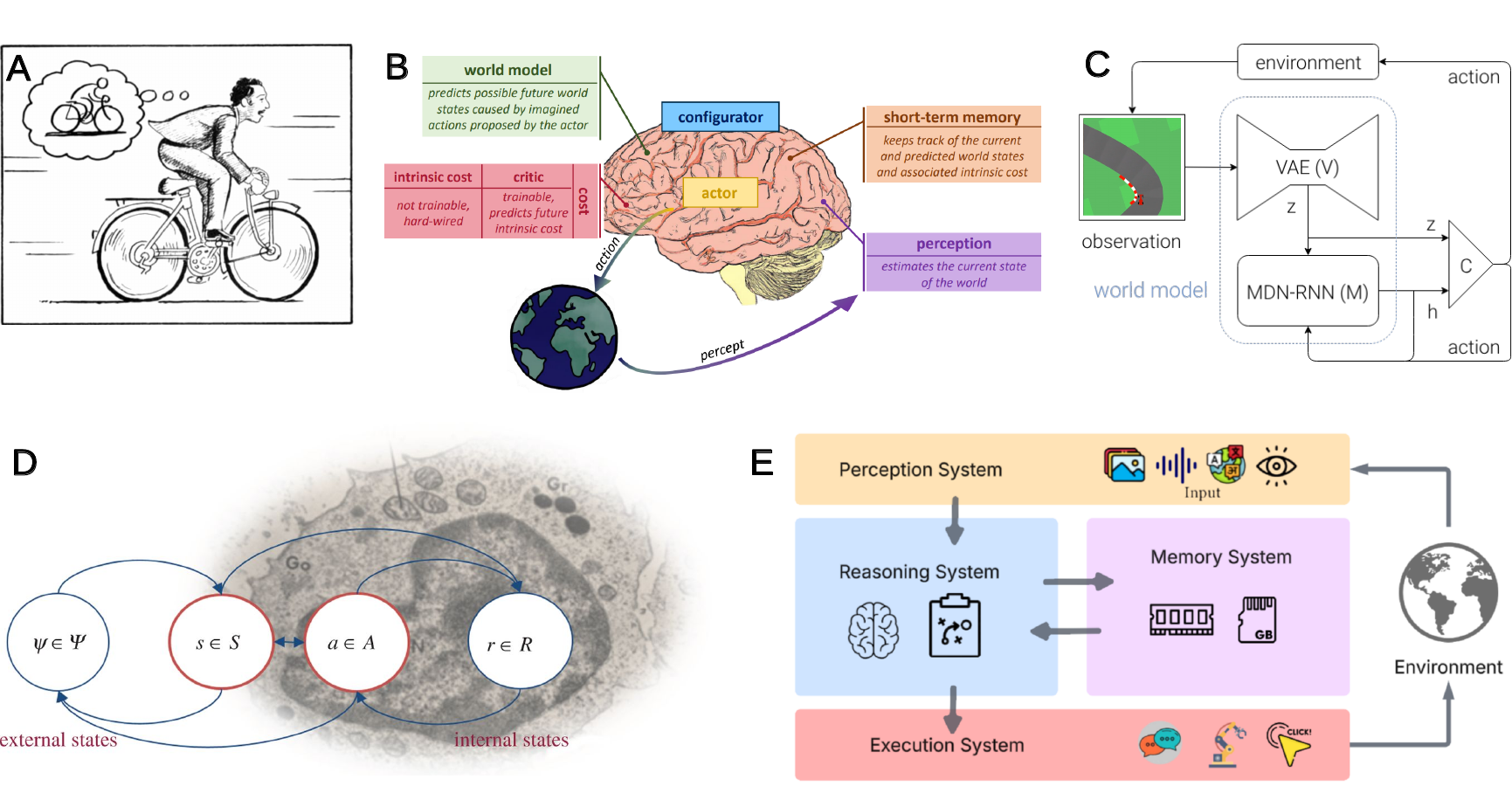}
    \caption{        
        \textbf{World models and agentic AI navigate the world via internal models.}
        (A) World model illustration of a cyclist from Scott McCloud’s Understanding Comics~\cite{mccloud1993understanding}; inspired by~\cite{ha2018worldmodels}.
        (B) A more novel and detailed illustration of the components of an agential world model architecture that learns to navigate a Joint-Embedding Predictive Architecture (JEPA); adapted from~\cite{dawid2023AutonomousMachineLearning}.
        (C) A possible (differentiable) world model implementation, comprising a perception encoder -- via a Variational Auto Encoder (VAE)~\cite{kingma2014auto} -- whose compressed latent representation $z$ is constantly contrasted to an internally predicted perception -- i.e., a predicted state of the world, generated by a Mixture Density Network (MDN) with a Recurrent Neural Network (RNN) backbone; both the compressed perception $z$ and internal (hidden) state $h$ of the MDN-RNN inform a controller module C that outputs an action with maximum return in a complex external environment, such as a car-racing track; adapted from~\cite{ha2018worldmodels}.
        (D) A schematic illustration of an active inference agent, partitioning the world into internal states and hidden or external states that are separated by a Markov blanket, comprising sensory and active states. Perception corresponds to the self-organization of internal states, while action couples internal states back to external states; adapted from \cite{Friston2015KnowingOnesPlace} and adapted from \cite{fields2022competency}. Through this perspective, active inference closely parallels world model architectures~\cite{Friston2021worldmodel}.
        (E) A relatively novel trend in large language model (LLM)-based reasoning systems integrates pretrained multimodal foundation models with external memory systems and tools for building ``Autonomous LLM Agents'' that can navigate custom problems actively by contrasting their understanding of the task (e.g., specified loosely via text prompts) to the context of an external world (such as accumulated experimental outcome and knowledge about the problem); adapted from~\cite{deLamo2025LLMAgents}.
        All of these architectures contrast an internalized representation, such as an embedding space of a (their) world to form decisions about optimal navigation policies in their environment, and/or remapping their internal representations based on novel evidence to execute better-informed decisions or actions in the future.
    }
    \label{fig:world-models}
\end{figure}

{\color{teal}Another recent promising trend in LLM–based reasoning systems integrates pretrained multimodal foundation models with external memory systems, tool-use interfaces, and autonomous control frameworks to build LLM agents capable of actively navigating custom problem domains~\cite{Wang2024SurveyLLMAgents, Park2023GenerativeAgents}. These systems dynamically contrast their internal task representations -- transforming a text-prompt specifying the problem into learned internal embedding representations -- with the unfolding state of an external world (e.g., accumulated experimental results, retrieved knowledge, or tool outputs) through mechanisms such as retrieval-augmented generation \cite{Lewis2020RAG, Borgeaud2022RAG}, tool-augmented action loops~\cite{Yao2023Tools, Mialon2023AugmentedLLMs2023}, and self-reflective or self-corrective reasoning procedures~\cite{Kojima2022ZeroShotLLMs, Shinn2023ReflexionLLMs, zhao2025learningreasonexternalrewards}. In this sense, emerging agentic LLM architectures operationalize the same twofold invariants we describe in this contribution: (i) remapping their embedding spaces as new context arrives -- via attention-based latent updates, external-memory retrieval, and world-model–like state revisions -- and (ii) navigating these evolving spaces through iterative error minimization, informed by planning, and action selection \cite{Huang2022WorldReasoningLLMs, Li2023EmergentWorldReprLLMs, lindsey2025biologyLLMs}. These systems increasingly resemble multiscale cognitive agents that refine internal latent spaces during inference, a property shared with biological navigation in embedding spaces.
LLMs thus represent an intriguing architecture that allow us to query the collective knowledge on the internet with incredible efficiently. 
The challenge might have shifted towards integrating those models into multi-scale architectures that can actively, and faithfully learn to understand and navigate relevant aspects of the world (see \Cref{fig:world-models} for an illustration).
A thorough understanding of multimodal embedding space geometries~\cite{Jha2025UniversalEmbedding}, and learning paradigms that lead to compositional~\cite{Kumar2025Fractured} or predictive multimodal (joint) embeddings ~\cite{dawid2023AutonomousMachineLearning} are promising directions for advancing the field of world model like architectures~\cite{ha2018worldmodels, Friston2021worldmodel} toward bio-inspired multiscale problem solving systems~\cite{levin2023darwin, hartl2025generativegenome}.}

Diffusion models (DMs) \cite{dhariwal2021DMBeatGANs, nichol2021improveddenoising, song2021denoising, ho2020denoising, sohl2015diffusionmodels}, on the other hand, are perhaps the clearest implementation of generative models that are intrinsically based on self-regulatory error-correction dynamics during both training and inference.  Rather than generating data in one deterministic pass, diffusion models are specifically trained to denoise potentially corrupted data: while a forward process successively corrupts data by incrementally adding noise to target data points, these models learn the reverse process of correcting noisy data samples. In that way, diffusion models not only learn to counter potential entropic -- or diffusive -- effects to restore relevant information, but represent powerful generative models that transform initial random noise into novel structured data samples that conform to the training data distribution. The generative process of diffusion models thus implements iterative self-regulatory error-correction dynamics, where each denoising step performs an incremental corrections on noisy data samples, gradually reconstructing complex structure from initially high-entropy states. 

This simple principle is the basis for cutting-edge image \cite{rombach2022high} or video generation techniques \cite{brooks2024videoworldsimulators}, and has also found application in generative game-play through world modeling \cite{alonso2024diffusion}, as backbone in optimization methods \cite{yan2024EmoDM, krishnamoorthy2023diffusionmodelsblackboxoptimization, zhang2025diffusionevolution, hartl2024hades}, but also in biomedically relevant applications such as \textit{in-silico} protein-folding \cite{Jumper2021} or protein-synthesis \cite{Watson2023} and as foundational model for genomics~\cite{Kenneweg2024SynteticGenotypesUsingDMs, Li2023LatentDMForDNA, Avdeyev2023DirichletDMsSequence}, transcriptomics~\cite{Zhang2025cfDiffusion, DaSilva2024DNADiffusion, Li2024stDiffusion, Luo2024scDiffusion, Lopez2018DeepGenerativeSCVAE}, etc.~\cite{Guo2023}.

\begin{figure}
    \centering
    \includegraphics[width=0.6\linewidth]{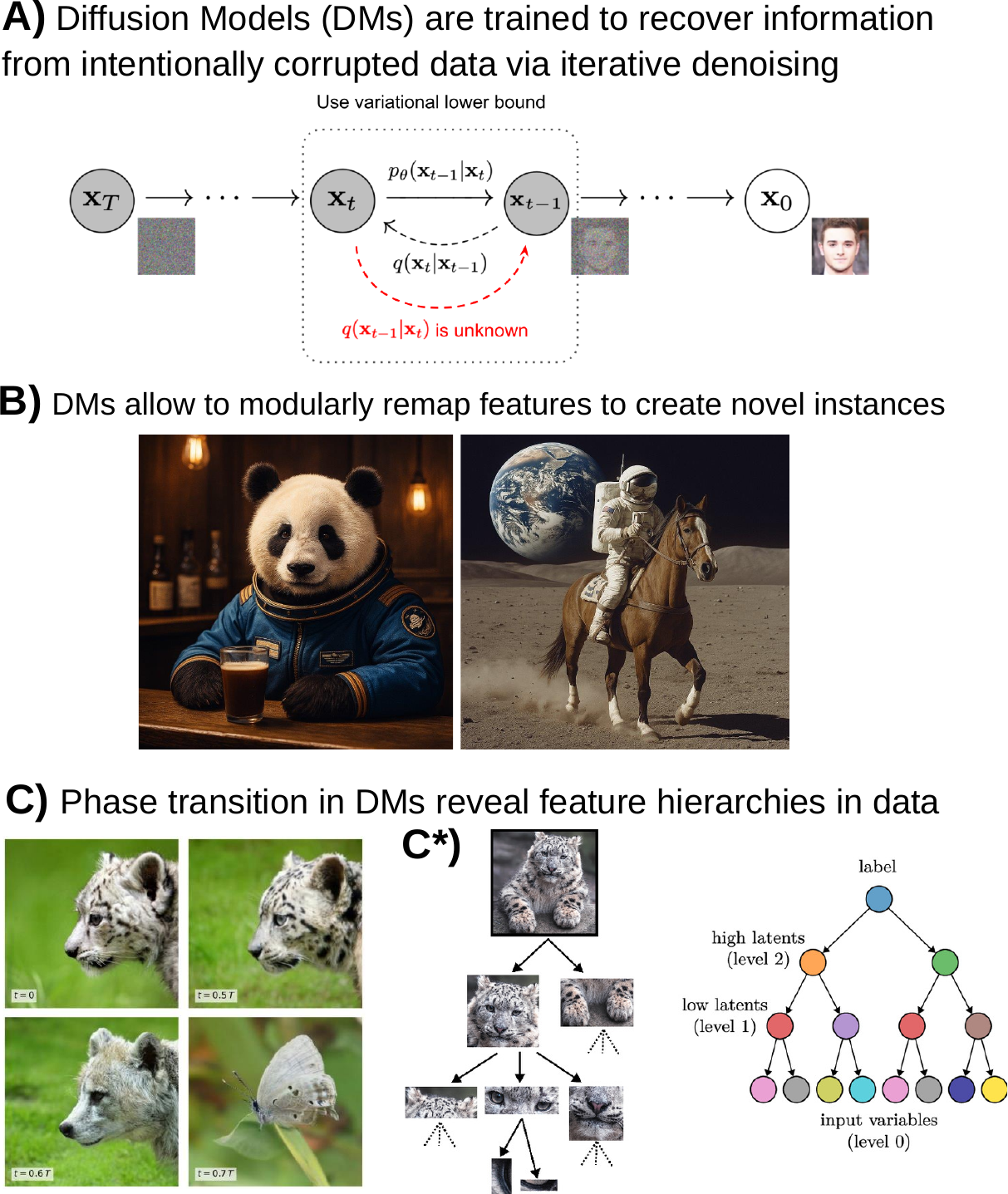}
    \caption{
        \textbf{Principles of diffusion models:}
        (A) In denoising diffusion models (DMs), a forward process corrupts the data from a particular domain (here the data is represented by an image) by incrementally blending it with noise. An ANN learns a reverse process that inverts this corruption process by predicting the noise component at each step. In that way, DMs learn the underlying probability distribution of unstructured datasets and have become state-of-the-art generative models for a wide range of applications.
        Novel data is generated by an iterative error-minimization process that induces a trajectory through latent space: the model navigates from high-entropy states toward attractor regions conforming to the data distribution. Along this trajectory, structured features successively emerge from noise through spontaneous symmetry breaking into feature-specific attractor states~\cite{Raya2023} (image adapted from \cite{ho2020denoising}). 
        (B) In that way, DMs can modularly recombine features across hierarchical levels, enabling compositional generation of novel samples not explicitly present in the training data;
        panel B shows AI-generated schematic illustration created using ChatGPT (OpenAI) based on author-provided text prompts.
        (C) DMs reveal the hierarchical nature of data: at a characteristic noise level, high-level semantic attributes (e.g., class identity) change abruptly, while low-level features vary smoothly across the entire process (c.f., species vs. facial details in the left panels of C). This reflects a layered decomposition of latent representations (C$^*$) that can be integrated by the DM at different stages during denoising (panels C and C$^*$ are adapted from \cite{Sclocchi2025}).
    }
    \label{fig:diffusion-models}
\end{figure}

\begin{figure}
    \centering
    \includegraphics[width=\linewidth]{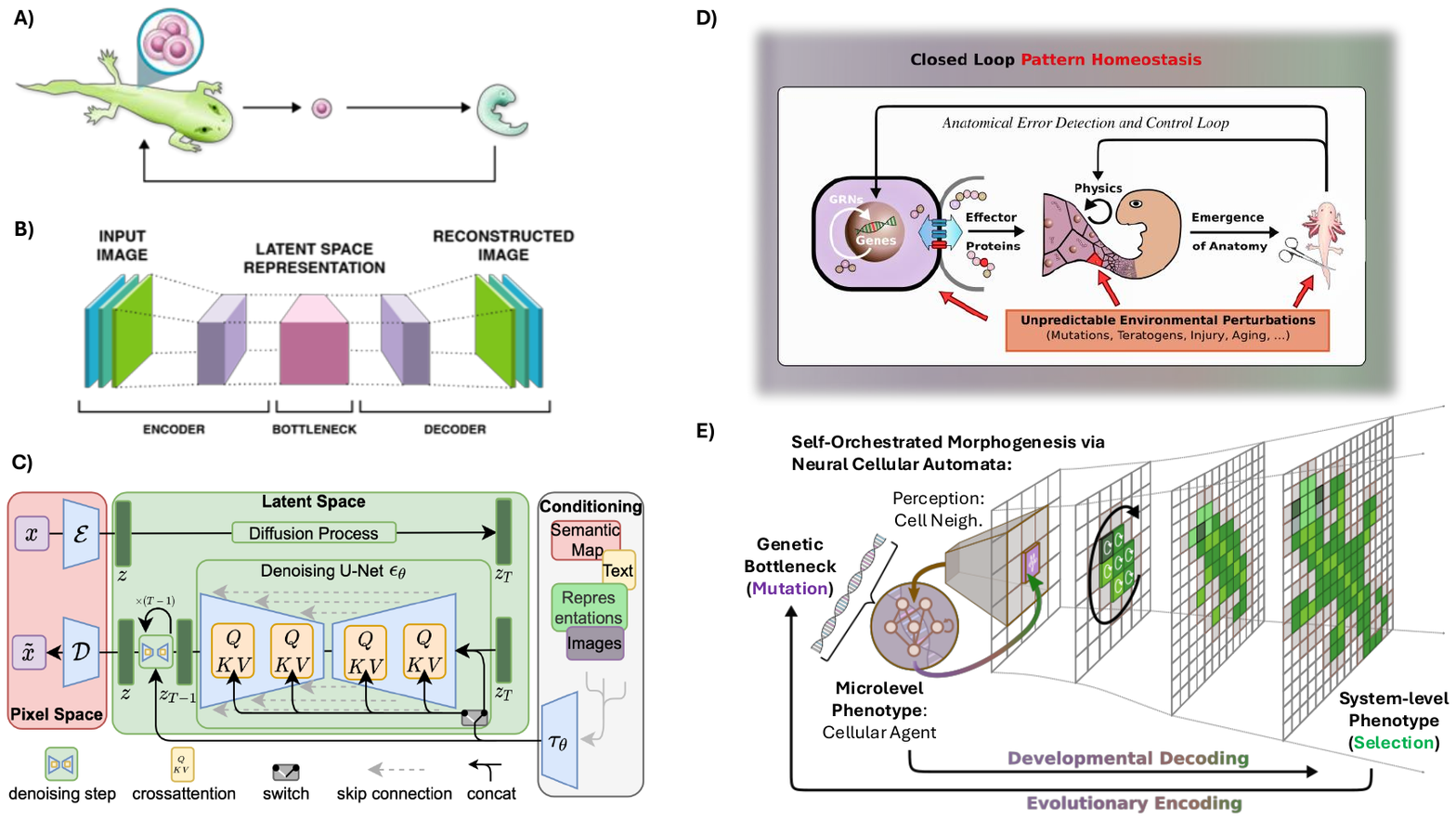}
    \caption{
        \textbf{Examples for navigating embedding spaces in unconventional substrates}:
        (\textbf{A}) Classical evolutionary development illustrates the canonical bow-tie organization of biological systems: a compact genomic bottleneck stores lineage-level regularities as latent variables~\cite{hartl2025generativegenome, mitchell2024genomic}, which are then decoded by the developing embryo into a coherent, species-specific anatomical phenotype. This process is not a rigid mapping but a context-sensitive, distributed reconstruction carried out by competent cellular agents navigating physiological and anatomical spaces.
        (\textbf{B}) Conceptually, this is paralleled in ML research by autoencoders (AEs), which similarly compress high-dimensional inputs into a low-dimensional bottleneck whose latent variables support modular recombination and high-fidelity reconstruction (and recombination). This provides a simplified example of how compressed variables can encode generative structure, though biological decoding is far more plastic, recurrent, and iteratively error-corrective than the single-shot downstream decoding of typical AEs.
        (\textbf{C}) Latent diffusion models exemplify a richer generative process: initial high-entropy samples are iteratively refined through successive denoising steps toward complex, structured attractors (c.f., \Cref{fig:diffusion-models}). These architectures embody incremental error correction, hierarchical feature emergence, and context-sensitive remapping, paralleling developmental morphogenesis~\cite{hartl2025generativegenome}, in which intermediate states are dynamically refined and recombined toward target anatomical outcomes.
        (\textbf{D}) Biological pattern homeostasis operates via nested closed-loop control systems: sub-cellular, cellular, tissue, and organ-level agents continuously detect deviations from preferred anatomical states and act to maintain homeostasis. These nested error-correction loops define a hierarchy of internal ``embedding spaces'' through which subsystems navigate, while higher levels reshape the energy landscape that lower-level agents inhabit, implementing the multiscale competency architecture of biology.
        (\textbf{E}) Neural cellular automata (NCAs) provide a synthetic analogue of this architecture. Each cell-like unit updates its internal state based on local perception-action loops (mediated by cell-specific artificial neural networks), and the collective of cells iteratively constructs a target morphology from a compact initial representation~\cite{hartl2025generativegenome, Hartl2025NCAs}. NCAs demonstrate how compressed encodings, combined with distributed, recurrent refinement protocols, can yield generative navigation protocols of cellular collectives toward system-level outcomes, reminiscent of the biological multiscale dynamics shown in (D).
        Images (A, B) by Jeremy Guay of Peregrine Creative used by permission from \cite{levin2023darwin}; 
        (C) adapted from \cite{rombach2022high};
        (D, E) used by permission from \cite{hartl2025generativegenome}.
    }
    \label{fig:architecture}
\end{figure}

Interestingly, this process can also be interpreted as an incremental teleologic process~\cite{zhang2025diffusionevolution}: while the system's dynamics executes small corrective steps (by design), the proposed state-corrections of the model are always with respect to a goal state represented by attractor states in the model's representation of the data manifold (embedding space) that are most similar to the current data pattern. Strikingly, this positions diffusion models as associative memories \cite{ambrogioni2024, krotov2025modernmethodsassociativememory} which undergo successive steps of symmetry breaking and phase transitions during their generative phase \cite{Raya2023, Ambrogioni2025, Sclocchi2025}.
During inference, diffusion models thus implement entropy-reducing generative trajectories that follow a hierarchy of self-regulated phase transitions through a learned latent space of a data manifold. We emphasize that this not only resembles but actually models biological processes in organismal development and cognition, based on self-regulatory error-correction principles in dynamic embedding spaces \cite{hartl2025generativegenome} (see \Cref{fig:diffusion-models}). This realization has allowed us to relate diffusion models with the process of evolution \cite{zhang2025diffusionevolution, hartl2024hades} and morphogenesis \cite{Hartl2024MCA}, explicitly framing the genome as a functional unit and literally as a generative model \cite{hartl2025generativegenome, mitchell2024genomic, pezzulo2016top} (see \Cref{fig:architecture}).

Morphogenesis describes the self-regulated growth of coherent anatomical structures of organisms. This describes trajectories through morphospace that start from high-entropic embryonic states, and which are then gradually refined into a mature morphology, rich in detail and structure. Importantly, this refinement happens gradually and is -- by far -- not deterministic. In contrast, morphogenesis relies on system-level feedback to gene regulatory dynamics at the subcellular level, and constructs organisms on the fly based on their current environmental context \cite{levin2023darwin}. For instance, even after significant injuries or perturbations, organisms such as planaria or amphibians can restore limbs, organs, or entire parts of their body~\cite{levin2023darwin}. Both morphogenesis and regenerative capacity imply the presence of internal models or reference frames of a target morphology (suitable for the current context~\cite{levin2021bioelectric, durant2017long}) and mechanisms of the cellular collective capable of detecting and correcting deviations of disturbed states back toward viable attractor morphologies. The agential biological substrate (especially at the cellular and subcellular levels) thus constantly needs to counteract entropic affects and diffusive processes conditional to their current environmental context and genetic prompts, not only to maintain their own physiological integrity, but to actively integrate into and maintain the collective multiscale competency architecture of the body \cite{hartl2025generativegenome, mitchell2024genomic}.

We thus argue that the training paradigm of diffusion models -- incremental denoising of entropic corruptions that can be susceptibly to external conditions \cite{ho2021classifierfree, rombach2022high} -- represents a strong inductive bias to the generative process of diffusion models that is based on non-equilibrium thermodynamics  \cite{sohl2015diffusionmodels} that perfectly aligns with biological organization principles \cite{friston2010, friston2016activeinference}.

At this stage, we would also like to emphasize similarities between diffusion models and transformers: despite having different architectures, both systems rely on incremental refinements of data representations.
In transformers, each layer incrementally refines the context of token embeddings based on skip-connections between attention maps. In diffusion models, each denoising step incrementally reduces estimated errors from noisy data samples and guides it toward a configuration that conforms to the training data distribution. 
Both processes thus implement a navigation policy in embedding spaces, governed by incremental refinements at each layer or timestep. Recently, both diffusion models and transformers have been associated to modern Hopfield-type associative memories \cite{ramsauer2021hopfield, ambrogioni2024, krotov2025modernmethodsassociativememory}, which moreover possess error-correction capabilities on the system-level scale, i.e., in restoring incomplete data and generating novel instances by recombining features of learned data modularly and creatively, a connection that becomes even more apparent with the novel ``Energy Transformer'' architecture \cite{hoover2023energy}. This parallel underscores how modern AI architectures increasingly align with principles of biological organization and cognition, where memory, planning, and perception all involve cycles of modeling, prediction, perception, and incremental error correction.

A closely related type of architecture to diffusion models with even more parallels to biological self-organization is that of Neural Cellular Automata (NCAs) \cite{Li2002, mordvintsev2022growing}.
Cellular Automata (CAs) are early models of \textit{Artificial Life} \cite{Langton1997, Langton1986} which date back to von Neumann's self-replicating machines~\cite{vonNeumann1966}. CAs have provided a foundational framework for modeling distributed biological phenomena such as replication, growth, and morphogenesis, simply through local interactions of neighboring cells on a discrete spatial grid. While CAs show life-like behavior \cite{gardner1970life, Wolfram2002}, tuning their hard-coded local update rules by hand proved to be difficult if not infeasible. A recent extension of CAs termed NCAs replaces their update rule with flexible and trainable neural networks, notable the same network per cell. Analogous to CAs, the dynamics of NCAs resides around cells updating their internal state based on local measurements of their neighboring cells' states. Strikingly, such NCAs can learn via differentiable~\cite{Randazzo2023, Mordvintsev2022IsotropicNCAs, Cavuoti2022, Randazzo2021, Niklasson2021, Randazzo2020, mordvintsev_growing_2020} or evolutionary learning frameworks \cite{montero_meta-learning_2024, Hartl2024MCA, lopez2023goals} the inverse problem of morphogenesis to grow a target morphology from a single seed cell. To name but a few more examples, NCAs can facilitate self-orchestrated computational tasks such as collective classification~\cite{Randazzo2020}, path-finding in a maze akin to slime-molds~\cite{earle2023pathfindingneuralcellularautomata, umu1729_maze_solver_2023}, and even decentralized control of composite robots \cite{Hartl2024MS, Sudhakaran2021, Risi2022a, najarro2022hypernca, Pathak2019},  all purely based on local intercellular communication and intracellular decision-making. NCAs can learn to regenerate complex patterns after damage or perturbation, reflecting a distributed capacity for self-regulation and repair, and might even allow us to understand and resolve aging \cite{PioLopez2025}; see \cite{Hartl2025NCAs} for a more comprehensive review. 

These models inherently operate as integrated networks of local agents continuously communication with each other via self-refinement, blurring the line between morphological development and information processing \cite{Levin2022TAME}. Although the exact training paradigms deviate, NCAs' dynamics reflect iterative self-regulatory (error-correcting) principles akin to the generative process of diffusion models:
an NCA's cells perceive their local environment, perform local cooperative communication, and iterative self-refinement until a collective target state is reached. By harnessing these principles, NCAs offer a powerful framework for understanding and engineering complex systems that need to reliably build and rebuild their form and function at the system-level from the bottom up; for a more in-depth discussion see \cite{hartl2025generativegenome}. Although trainability and scalability challenges remain\cite{bielawski_evolving_2024, Pande2023, palm2022variational}, this makes NCAs powerful tools for encoding and navigating embedding spaces in a bio-akin way, where localized deviations must be corrected while preserving global structure, a property highly reminiscent of tissue-level repair in biology.

Living systems are organized into multiscale competency architectures \cite{levin2023darwin}. At different hierarchical levels -- molecular, cellular, tissue, organism -- agents possess goals and the means to achieve those goal states via local error correction mechanisms: Cells can adjust gene expression profiles, tissues can reconfigure morphogen gradients, entire organisms can reroute behaviors, etc. The actors at each scale thus navigate their own embedding spaces, while contributing to the stability and adaptability of the system as a whole and scales into an entire problem-solving layer between the genome and phenotype of an organism \cite{levin2023darwin}. Furthermore, such a nested architecture enables concurrent processes to occur simultaneously at different scales and renders biology as an inherently polycomputing substrate \cite{bongard_2023_plentyofroom}. This allows biological systems to process multiple streams of information in parallel and solve multiple problems simultaneously and across scales: the hardware of one layer may become the signals of yet another layer of organization.

Altogether, these examples suggest that the fundamental organizational principle underlying both biological morphogenesis, cognitive processes, and modern AI systems like transformers, diffusion models, and NCAs is a form of remapping and navigating embedding spaces through error correction. Whether generating a coherent sentence, synthesizing a photorealistic image, or assembling a limb, self-regulative systems leverage iterative cycles of prediction and refinement to guide patterns through embedding spaces. In turn, this suggests that integrative (scale-free) error correction may be the essential computational backbone of intelligence itself, enabling systems to maintain their integrity while remaining flexible enough to adapt, innovate, and explore. All intelligence is collective \cite{mcmillen2024collective, Levin2025DiverseIntelligence}, and the best localized agents can do depends on the extent of their cognitive light cone \cite{Levin2019}. Collective error correction across a nested hierarchy of agents \cite{levin2023darwin} might leverage self-regularized criticality \cite{HENGEN2025, dunkelmann1994neural, Hesse2014, Beggs2014} as an operational setpoint for scaling intelligence across scales.
 
\section{Implications, Connections, Broader links}
% \begin{itemize}
%     \item Meaning for AI/AGI research (how do new levels of cognition emerge? Is global coordination under local competition enough?) 
%     \item Everything can be maybe defined as a Hopfield net (even in the modern Hopfield network \cite{krotov2025modernmethodsassociativememory}, the generative process is a single energy-minimization step, i.e., a single error correction mechanism in an energy embedding space) 
%     \item "agents", "generative models", and "non-equilibrium thermodynamics"? 
%     \item What does it mean if we are right – what can you do? What will future look like? 
%     \item {\color{teal} We should also mention the ideas from Monday's lab meeting presentation (Sept. 15, 2025, by Jia Liu) on \textit{Drift to Remember} dynamics \cite{Du2024DritToRemember} -> extending attractor landscapes (in embedding spaces) as a mechanism for life-long learning. [maybe also related to JEPA; and potentially a property of transformers' associative memory, which might be why they don't suffer catastrophic forgetting?]}
%     \item We might also want talk about AI agents, they are also navigating multimodal space, and have tool-use, see https://arxiv.org/pdf/2510.09244
% \end{itemize}

% BELOW IS A DRAFT FOR THE "IMPLICATIONS, CONNECTIONS, BROADER LINKS" SECTION - citations will follow
Identifying the interplay of remapping and navigation within embedding spaces as invariant of cognition provides a unifying lens on adaptive intelligence across substrates. Integrated biological systems not only embody, but continually construct and revise internal maps -- from transcriptional states to morphogenetic fields, organoids, organisms, and beyond -- and use distributed error correction mechanisms toward stable attractor states at the (collective) system level. AI systems, though implemented in very different substrates, increasingly exploit the same logic: embeddings compress complex data manifolds into navigable spaces, and iterative refinement steps drive trajectories toward coherent, increasingly contextualized outcomes. This dual principle positions cognition not as a property of a particular substrate, i.e., biological matter, but as a transformative process of continual (active) model refinement under compositional error minimization towards a system-level agenda.

One implication of this framework is that intelligence becomes inherently collective and multiscale -- actors at one level can become the signal at other levels. In biology, localized agents, such as gene regulatory networks or cells, act in their own embedding spaces while higher-level systems reshape the energy landscapes of their parts. When integrated accordingly, this architecture enables robust scaling from molecular to organismal competencies. In turn, collective AI systems may prove more adaptive than monolithic models: ensembles of agents can maintain partially private embeddings and must learn to generalize from constrained environmental information, yet participate in shared embedding spaces, coordinating through distributed error correction. Such architectures align with proposals that advanced intelligent systems must develop shared world models with humans to form collaborative systems, reframing alignment as a synergetic co-remapping (social) process rather than a one-sided constraint~\cite{pezzulo2025sharedworlds}.

Second, the dual principle of remapping and navigation underscores scale-free incremental error correction as the operational backbone of cognition. Both biological organisms and modern AI models succeed not by providing predictions or actions in one pass, but by iteratively detecting mismatches against projected perceptions and corresponding refinements to the target state based on internal navigation policies. Importantly, many systems use the same, or similar mechanisms for learning and inference: the rules that remap embeddings during training also govern navigation protocols at runtime. These systems thus essentially learn transformative error-correction processes in diverse loss landscapes rather than attractor states directly, which represents a strong inductive, bio-inspired bias toward self-organizing systems that are robust against entropic disorder. This suggests that future intelligent systems should unify learning and deployment under a single engine of iterative refinement in dynamically remappable embedding spaces.

Finally, we propose {\color{teal} near-}criticality as a candidate universal setpoint for such scale-free, substrate agnostic cognitive engines~\cite{HENGEN2025, fields2020scale}. Systems near critical regimes exhibit the very properties required for continual remapping {\color{teal} --} sensitivity to perturbation, scale invariance, tunability, and maximal information throughput{\color{teal}: Perturbations or novel contexts can transiently bring parts of such systems toward critical regimes, where the (sub-)space of alternative states becomes statistically accessible by remapping dynamics. Weak cues can rapidly bias the system toward particular attractors, breaking the symmetries among competing alternatives.} Evidence from neuroscience suggests that brains exploit such near-critical dynamics~\cite{dunkelmann1994neural, Hesse2014, Beggs2014}, and theoretical work argues that criticality may be the natural operating point for adaptive systems~\cite{HENGEN2025}. Yet criticality is fragile: small fluctuations can cascade, and training artificial networks close to criticality can destabilize learning. For future self-regulatory AI systems, this suggests a delicate balance: architectures must incorporate homeostatic mechanisms to navigate toward stable attractors, yet must be able to exploit critical regimes for adaptability if necessary. Criticality might be a substrate independent natural fallback mechanism of collective self-regulatory systems that explicitly enables exploration and search upon perturbation {\color{teal}or under stress}, but which must be actively regulated otherwise to allow stable navigation toward viable attractor states.
{\color{teal} Being in a near critical state does not simply ``cause'' remapping, nor do remapping systems automatically become critical; rather, adaptive biological systems may have evolved (because of) regulatory mechanisms that transiently bring subsystems toward near-critical regimes under perturbation, where large-scale remapping of internal embedding spaces is most effective, before homeostatic mechanisms restore stable navigation within an updated system-level attractor landscape.}

In conclusion, we frame remapping and navigation of embedding spaces as a scale-free organizational principle of cognition. This dual process links developmental robustness with generative modeling, relates collective AI within the same framework as multicellular coordination, and suggests (near-)criticality as a dynamical setpoint naturally supporting the required balance between stability and adaptability in cognitive systems.  To the general framework of the Free-Energy Principle \cite{ramstead:22, friston:23}, it adds architectural principles with which the generative models of systems with sufficient degrees of freedom and internal structure -- including living systems at all scales -- can be expected to comply. By situating remapping and navigation as invariants of cognition, we suggest a substrate-independent principle that links diverse intelligences across scales. This perspective emphasizes not just descriptive parallels, but also operational mechanisms for how adaptive and resilient minds -- biological, artificial, or hybrid -- emerge from shared (self-)organizational dynamics.
    
%%% CONTENT %%%%%%%%%%%%%%%%%%%%%%%%%%%%%%%
%%%%%%%%%%%%%%%%%%%%%%%%%%%%%%%%%%%%%%%%%%%

\newpage
\section*{Figures}
\FloatBarrier

% NOTES FROM TOMIKA:

% swap , costs 150$
% - symmetry breaking
% - remove

% Fig. 4c -> not mixing (either use originals - not mix)

% fig. 2b -> make own version ...

% fig. 3b -> use other one !!!!

% fig. 5c -> 16$ vs. 68$ | if someone is IEEE member

% ---

% would need pay twice

% \begin{figure}
%     \centering
%     \includegraphics[width=\linewidth]{media/Figure-5-Platonic-Representation.pdf}
%     \caption{
%         {\color{red}[DRAFT - TODO]}
%         \textbf{Embedding spaces in AI converge towards platonic representations.}
%         (A) Embedding models of different data modalities (text, image, sound) find the same semantic embedding representation, with conserved distance metrics between the same concepts \cite{Huh2024PlatonicRepresentation, yang2024harnessing}.
%         (B) Different training procedures in ANNs after can lead to conceptually different internal representations either capable of feature separation (left, neuroevolution) or with entangled feature representations (right, backpropagation) \cite{Kumar2025Fractured}.
%         Open-ended evolution might favor disentangled representations in favor of modular evolvability.
%     }
%     \label{fig:platonic}
% \end{figure}

%%%%%%%%%%%%%%%%%%%%%%%%%%%%%%%%%%%%%%%%%%%
%%% BACK MATTER %%%%%%%%%%%%%%%%%%%%%%%%%%%
\newpage
\section*{Acknowledgements}

%%%  Use this section to acknowledge contributions 
%%%  from non-authors and list funding sources, 
%%%  including grant numbers.
We thank members of the Levin Lab for helpful discussions, and Tomika Gotch for help with the manuscript. 
We acknowledge support from Astonishing Labs, Inc., and the Templeton World Charity Foundation, Inc. (TWCF0606).
This publication was made possible through the support of Grant 62212 from the John Templeton Foundation. 
The opinions expressed in this publication are those of the author(s) and do not necessarily reflect the views of the John Templeton Foundation.

%%%%%%%%%%%%%%%%%%%%%%%%%%%%%%%%%%%%%%%%%%%
%%% REFERENCES %%%%%%%%%%%%%%%%%%%%%%%%%%%%
\bibliography{references}
\bibliographystyle{unsrtnat}

%%% REFERENCES %%%%%%%%%%%%%%%%%%%%%%%%%%%%
%%%%%%%%%%%%%%%%%%%%%%%%%%%%%%%%%%%%%%%%%%%

%%% BACK MATTER %%%%%%%%%%%%%%%%%%%%%%%%%%%
%%%%%%%%%%%%%%%%%%%%%%%%%%%%%%%%%%%%%%%%%%%
\end{document}